\pdfoutput=1

\documentclass[10pt,twocolumn,letterpaper]{article}

\usepackage[pagenumbers]{cvpr}

\usepackage{cuted} 

\usepackage{currfile} 

\usepackage{caption} 
\usepackage{booktabs} 
\usepackage{multirow} 
\usepackage{placeins} 
\usepackage{float} 
 \usepackage{fancyhdr}

\definecolor{cvprblue}{rgb}{0.21,0.49,0.74}
\usepackage[breaklinks,colorlinks,allcolors=cvprblue]{hyperref}
\hypersetup{
  pdftitle={LEF-JEPA: Denoising-Trained Latent Energy Fields for Spatial Representation Learning},
  pdfauthor={Zhiyuan Liu; Zhihao Xu; Sicheng Liu; Huanhuan Huang}
}

\def\confName{CVPR}
\def\confYear{2027}

\title{LEF-JEPA: Denoising-Trained Latent Energy Fields for Spatial Representation Learning}

\author{%
  {\large Zhiyuan Liu\textsuperscript{1,2,*,\textdagger} \quad
  Zhihao Xu\textsuperscript{1,*} \quad
  Sicheng Liu\textsuperscript{1} \quad
  Huanhuan Huang\textsuperscript{1}}\\[0.35em]
  {\normalsize \textsuperscript{1}Qinous Technology Co., Ltd \quad
  \textsuperscript{2}Tsinghua University}
}

\newcommand{\paperrunningtitle}{LEF-JEPA: Denoising-Trained Latent Energy Fields for Spatial Representation Learning}
\fancypagestyle{firstpage}{%
  \fancyhf{}%
  \fancyfoot[C]{\thepage}%
}

\makeatletter
\renewcommand{\@maketitle}{%
  \newpage
  \null
  \vspace*{-0.20in}%
  \begin{center}
    \hrule height 1pt
    \vskip 0.25in
    {\Large\bfseries \@title\par}
    \vskip 0.22in
    \hrule height 1pt
    \vskip 0.24in
    {\@author\par}
    \vskip -0.25in
  \end{center}%
}
\makeatother

\begin{document}
\maketitle
\thispagestyle{firstpage}
\begingroup
\renewcommand{\thefootnote}{\fnsymbol{footnote}}
\footnotetext[1]{Equal contribution.}
\footnotetext[2]{\raggedright Corresponding author: Zhiyuan Liu\\[-0.15em]
  (\href{mailto:liuzhiyu22@mails.tsinghua.edu.cn}{\texttt{liuzhiyu22@mails.tsinghua.edu.cn}}).}
\endgroup
\begin{strip}
\centering
\includegraphics[width=0.90\linewidth]{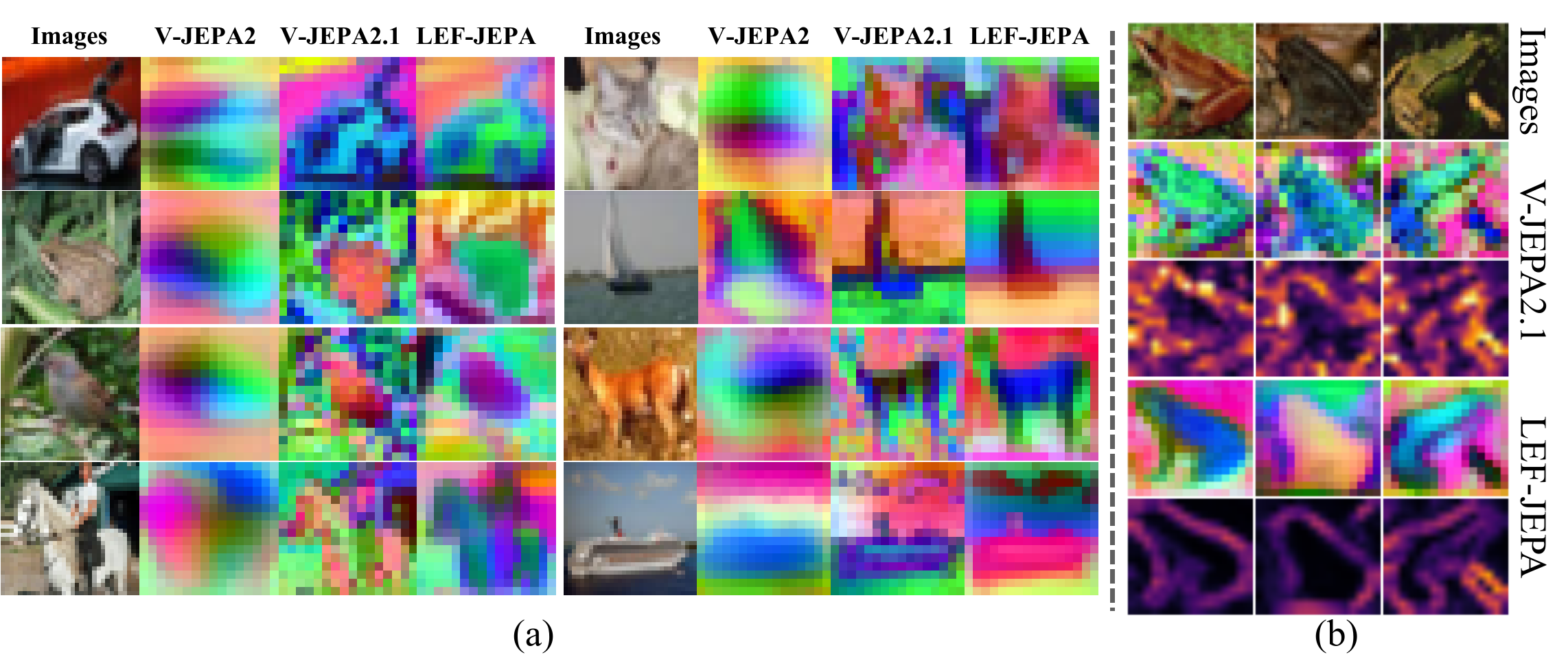}
\captionsetup{hypcap=false}
\captionof{figure}{\textbf{Recognizable object structure within coherent spatial regions.}
In the PCA visualizations, V-JEPA2 produces relatively coarse-grained
representations in which object shapes and boundaries are not clearly
delineated; V-JEPA2.1 recovers more local structure but also retains more
texture and color variation. In contrast, LEF-JEPA preserves recognizable
object shapes and boundaries while forming more coherent spatial regions (a).
Bright colors indicate larger
representational differences between adjacent patches in the PCA subspace,
whereas dark colors indicate locally smoother regions. Bright responses are
more broadly distributed for V-JEPA2.1, whereas for LEF-JEPA they concentrate
around contours and structural boundaries (b).}
\captionsetup{hypcap=true}
\label{fig:teaser}%
\end{strip}
 \begin{abstract}
A joint-embedding predictive architecture (JEPA) that applies prediction
supervision only at masked positions struggles to clearly capture object shapes
and boundaries in its latent representations. We further observe that some
existing methods improve spatial structure while also retaining rich variations
in texture and color. However, preserving more low-level visual information
does not necessarily make representations better suited to prediction and
action. This raises a key question: how can we strengthen the spatial structure
in latent representations that supports prediction and action without
over-encoding these details? We introduce LEF-JEPA, which learns spatial
structure through target-guided latent recovery and transfers this supervision
to the visual encoder through a relative-energy constraint; the energy module
is used only during pretraining and adds no inference overhead. To reduce the
cost of energy optimization during pretraining, we further construct a
conditional Ising Hamiltonian that allows low-energy-state search to be
performed by specialized Ising solvers. On physical hardware, we validate the
feasibility of achieving lower forward latency with fewer local updates. Under
a unified CIFAR-10 pretraining and frozen-encoder evaluation protocol, both
realizations outperform V-JEPA2.1 in image classification, foreground
segmentation, and visual planning.
\end{abstract}
 \section{Introduction}
\label{sec:intro}

Visual world models need to preserve the scene geometry required for prediction
and action. The Joint-Embedding Predictive Architecture (JEPA) uses features
extracted by an encoder as prediction targets without encoding every pixel-level
detail in the observation~\cite{lecun2022path,dawid2023latent}. As JEPA applications extend from
image-level recognition to pixel-level scene parsing and action-conditioned
planning~\cite{murlabadia2026vjepa21,assran2025vjepa2,terver2026ebjepa,nam2026causaljepa,wang2026temporal},
high-level semantics alone no longer suffice: segmentation requires clear
object boundaries, while planning depends on spatial layout and relations among
objects. Visual encoders must therefore preserve the local
spatial structure required for prediction and action while forming high-level
semantics.

However, when predictive supervision is applied only to masked positions, JEPA
struggles to clearly capture object shapes and boundaries in latent
representations~\cite{murlabadia2026vjepa21}. The V-JEPA2 features in
Figure~\ref{fig:teaser}(a) form coarse regions in which object contours and
internal parts are weakly distinguished. V-JEPA2.1 strengthens dense prediction
and multi-level self-supervision to improve the representation of spatial
structure~\cite{murlabadia2026vjepa21}. We further observe that this improvement
is accompanied by rich texture and color variations: the local-roughness
responses in Figure~\ref{fig:teaser}(b) are broadly distributed across object
interiors and backgrounds. Preserving more appearance detail does not
necessarily make representations better suited to prediction and action. The
key question is therefore how to strengthen the spatial structure in latent
representations that supports prediction and action without requiring detailed
encoding of low-level visual information.

We use spatial-structure cues in multi-level target representations to guide
the local recovery of perturbed latent representations. However, even if the
recovered representations capture these cues, the visual encoder's direct
outputs may not retain the corresponding spatial information. Because
downstream tasks directly use encoder representations, the supervision
provided by the recovery process must be
transferred to the encoder during pretraining, so that downstream tasks do not
depend on iterative recovery at test time~\cite{gladstone2026energy}.

Accordingly, we introduce LEF-JEPA, which augments masked prediction with
target-guided latent energy learning. It uses multi-level targets to learn
local recovery directions for perturbed representations and transfers this
structural supervision to the visual encoder through a relative-energy
constraint. The two training paths are decoupled, so encoder updates do not
backpropagate through the multi-step recovery process; the energy module is
used only during pretraining and therefore adds no downstream inference
overhead.

To explore more efficient energy learning during pretraining, we further
construct a conditional Ising Hamiltonian that enables LEF-JEPA to interface
with specialized Ising solvers. Unlike prior work that primarily uses physical
Ising devices for sampling in generative models~\cite{bohm2022noise,zhu2026kaiwu},
we incorporate physical low-energy-state search into self-supervised
pretraining and validate this closed-loop training process on a physical
photonic coherent Ising machine (CIM).

We conduct self-supervised pretraining on CIFAR-10~\cite{krizhevsky2009cifar} and freeze
the encoders for image classification, out-of-distribution transfer to
PartImageNet~\cite{he2022partimagenet} foreground segmentation, and
Two-Room visual planning. Across ViT-S and ViT-B, LEF-JEPA breaks the observed
tradeoff between semantic and spatial capability. The continuous neural field
achieves the strongest ViT-S foreground segmentation, whereas the Ising
realization solved on a physical CIM requires only a single local update step
($K=1$) and lower pretraining latency while attaining the strongest
classification and visual planning results at both scales, together with the
strongest ViT-B foreground segmentation.

The contributions of this work are:
\begin{itemize}[leftmargin=*,nosep]
  \item \textbf{Target-guided latent energy field:} Using structural cues from
  multi-level targets to learn local recovery directions for perturbed
  representations. The energy field is used only during pretraining and adds
  no inference overhead.
  \item \textbf{Decoupled dual-path energy training:} Sharing one energy
  function while updating the energy field and encoder separately. The encoder
  receives supervision through a one-sided relative-energy constraint without
  backpropagating through multi-step recovery.
  \item \textbf{Hardware-accelerated Ising realization:} Formulating a
  conditional Ising Hamiltonian compatible with specialized Ising quantum
  hardware and validating the closed training loop on a physical CIM, using a
  single update to accelerate and improve self-supervised representation
  learning.
\end{itemize}

 \section{Related Work}
\label{sec:related-work}

\subsection{Latent Prediction and Spatial Representation Learning}

I-JEPA uses visible regions within a single image to predict features of masked
regions~\cite{assran2023ijepa}; V-JEPA extends this task to video, using
spatiotemporal context to predict features of masked video
blocks~\cite{bardes2024vjepa}. V-JEPA2 extends this paradigm to an
action-conditioned world model for visual planning, establishing a common
foundation for recognition and control~\cite{assran2025vjepa2}. Causal-JEPA
uses object-level latent masking to learn interaction-dependent object
dynamics~\cite{nam2026causaljepa}, while Temporal Straightening constrains
latent-trajectory curvature to make the representation geometry better suited
to planning~\cite{wang2026temporal}. Together, these works show that
representations for prediction and action should not only preserve spatial
structure, but also emphasize information relevant to interactions and
dynamics rather than indiscriminately encoding all low-level visual information
in the observations.

Recent work designs training constraints around local features and their
spatial relations to improve performance on tasks such as segmentation and
depth estimation. V-JEPA2.1 supervises both visible and masked
tokens at multiple encoder depths~\cite{murlabadia2026vjepa21}. Because it
likewise uses multi-level targets but applies supervision through direct
regression, it provides the closest regression-based counterpart to the energy
formulation of LEF-JEPA. DINOv3, meanwhile, introduces Gram anchoring to
preserve pairwise similarities among local features over long training
schedules~\cite{simeoni2026dinov3}. Together, these approaches strengthen the
encoder's ability to extract structured spatial
information from pixel observations, showing that local spatial relations
benefit from constraints beyond standard masked prediction.
Another line of work regularizes global consistency and distributional shape
in latent representations. LeVJEPA
combines global--local video-view consistency with SIGReg, allowing
semantically organized regions to emerge even though the loss reads only
clip-level embeddings~\cite{kuhn2026levjepa}. VISReg decouples embedding scale
from distributional shape through variance and Sliced-Wasserstein constraints
~\cite{wu2026visreg}. Explicit geometric information can also serve as
pretraining supervision: LingBot-Vision uses online subpixel boundary detection
to guide masked prediction
~\cite{fu2026spatial}. Unlike explicit boundary detection or dense token
regression, LEF-JEPA uses multi-level target representations to supervise a
conditional energy field in learning local recovery directions for perturbed
representations, then transfers the supervision provided by recovery to the
visual encoder through a one-sided relative-energy constraint.

\subsection{Energy Learning and Physical Ising Solvers}

Energy-based models (EBMs) formulate inference and prediction as the search for
low-energy configurations~\cite{lecun2006tutorial}. This principle has been
applied to state-space trajectory planning~\cite{du2020planning}, token-level
associative-memory updates in Energy Transformer~\cite{hoover2023energy}, and
iterative test-time refinement of candidate predictions in Energy-Based
Transformers~\cite{gladstone2026energy}. Within JEPA-based representation learning, EB-JEPA
defines latent prediction error itself as energy and combines it with
VICReg~\cite{bardes2022vicreg} or SIGReg~\cite{balestriero2025lejepa} to avoid
representation collapse~\cite{terver2026ebjepa}. LEF-JEPA
does not redefine the prediction loss as energy or require iterative gradient
descent at inference. It retains standard masked prediction and introduces an
auxiliary conditional energy mechanism only during pretraining. The energy-field
objective learns local recovery directions guided by multi-level targets, while
the one-sided relative-energy constraint transfers this supervision to the
visual encoder without backpropagating through the multi-step recovery process
during encoder updates.

Beyond direct parameterization by continuous neural networks, conditional
energy landscapes can also be constructed with discrete spin models. The
central challenge is to search low-energy states efficiently and convert them
into training gradients. Neural Boltzmann Machines, for example, generate the
parameters of a restricted Boltzmann model from context and analytically
marginalize hidden units to obtain conditional free energy, yet still rely on
Markov chain Monte Carlo (MCMC) sampling on classical architectures to estimate
generative-likelihood gradients~\cite{lang2023neural}. To overcome the
efficiency bottleneck of software sampling, specialized physical Ising
solvers---including superconducting quantum annealers such as D-Wave that
exploit quantum tunneling~\cite{kadowaki1998quantum,johnson2011quantum} and
coherent Ising machines (CIMs) based on coupled optical oscillators
~\cite{inagaki2016cim,yamamoto2017cim}---use the natural evolution of physical
systems for high-speed search over low-energy spin configurations. Existing
machine-learning studies that employ physical Ising hardware, whether through
quantum-annealing sampling~\cite{adachi2015application,amin2018quantum} or
noise-injected optoelectronic sampling~\cite{bohm2022noise}, primarily treat it
as a Boltzmann sampler for estimating negative-phase statistics in generative
models and do not directly turn discrete hardware samples into gradients with
respect to continuous visual representations. LEF-JEPA instead extends physical Ising
solving from generative-likelihood estimation to self-supervised visual
representation learning. Its conditional Ising Hamiltonian incorporates
low-energy-state search into pretraining, allowing specialized Ising solvers to
participate in latent-space energy learning.
 \begin{figure*}[t]
  \centering
  \includegraphics[width=\textwidth]{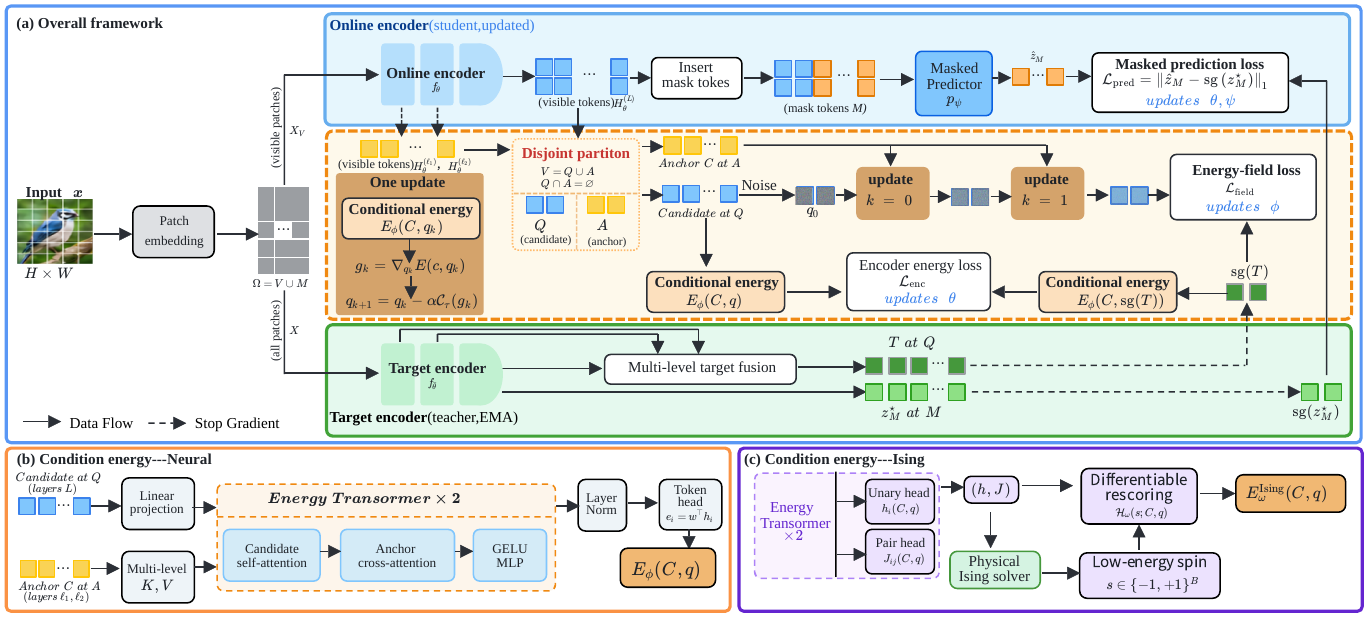}
  \caption{\textbf{Overview of LEF-JEPA.}
  (a) Visible tokens from the online encoder are partitioned into disjoint
  candidate and anchor sets. The anchors condition the latent energy landscape
  over perturbed candidate representations. Multi-level target representations
  both supervise the displacement induced by $K$ energy-gradient updates
  ($\mathcal L_{\mathrm{field}}$) and provide the target-energy reference for
  visual-encoder supervision ($\mathcal L_{\mathrm{enc}}$), together with the
  masked-prediction objective ($\mathcal L_{\mathrm{pred}}$). (b) The neural
  energy realization contextualizes candidate representations with multi-level
  anchors and aggregates the energy contribution of each token. (c) The
  conditional Ising realization predicts Hamiltonian coefficients $(h,J)$,
  invokes a physical photonic CIM to search for low-energy spin states, and
  differentiably rescores them to form a smooth scalar energy.}
  \label{fig:method-overview}
\end{figure*}

\section{Method}
\label{sec:method}

LEF-JEPA augments masked latent prediction with target-guided conditional
energy learning over visible tokens. To endow the visual encoder with spatial
coherence without iterative correction at inference, two complementary
processes are connected through a shared energy function. The Energy Field
Loss ($\mathcal L_{\mathrm{field}}$) supervises local correction displacements
of perturbed representations to shape the energy landscape. The Encoder Energy
Loss ($\mathcal L_{\mathrm{enc}}$) encourages online representations to lie
below the energy of a multi-level target reference, so that the learned
structure is reflected in encoder representations. The objectives share the
energy function, but their gradients follow strictly separated parameter paths.
Figure~\ref{fig:method-overview} summarizes the complete framework.

\subsection{Masked Latent Prediction}

Let $\Omega=V\cup M$ denote the visible and masked patch-position sets of image
$x$. The online encoder $f_\theta$ encodes the visible patches $x_V$, while the
masked predictor $p_\psi$ predicts latent representations at the masked
positions $M$. The target encoder $f_{\bar\theta}$ processes the unmasked view
to provide supervision; its parameters are updated by an exponential moving
average (EMA) of the online encoder and receive no gradients. With
$z_j^{\star}$ denoting a target patch token, the Masked Prediction Loss is
\begin{equation}
\mathcal L_{\mathrm{pred}}
=\frac{1}{|M|D}\sum_{j\in M}
\left\|p_\psi(f_\theta(x_V))_j-
\operatorname{sg}(z_j^{\star})\right\|_1
\label{eq:prediction}
\end{equation}

\subsection{Target-Guided Spatial Objects}

To model spatial relations among visible patches, we construct three
representations over $V$: candidate states $S$ to be corrected, contextual
conditions $C$, and a multi-level target reference $T$. The visible positions
are partitioned into disjoint candidate and anchor sets $Q$ and $A$, with
$V=Q\mathbin{\dot\cup}A$ and $Q\cap A=\varnothing$. These sets select $S$ and
$C$ from the online encoder and $T$ from the target encoder.

A central requirement is that the energy remain sensitive to relative spatial
variation across patches while keeping feature scale calibrated. Standard
token-wise normalization removes relative magnitude differences between object
boundaries and interior regions. We instead use contrast-preserving spatial
normalization. Given the $n$ patch tokens of one image,
$H=[h_1,\ldots,h_n]\in\mathbb R^{n\times D}$, we subtract the spatial mean
$\mu(H)$ to remove the component shared by the full image, then scale all
tokens by a single RMS factor computed jointly over positions and channels:
\begin{equation}
\mathcal N_s(H)=
\frac{H-\mu(H)}{
\sqrt{\frac{1}{nD}\|H-\mu(H)\|_F^2+\epsilon}},
\quad
\mu(H)=\frac{1}{n}\sum_j h_j
\label{eq:spatial-normalization}
\end{equation}
All positions share this RMS scalar, preserving their relative spatial
salience. For a transformer of depth $L$ with selected intermediate layers
$\ell_1,\ell_2$, the clean online candidate source and multi-level anchor
condition are
\begin{equation}
S=\mathcal N_s(H_\theta^{(L)})_Q,\qquad
C=\left\{
\mathcal N_s(H_\theta^{(\ell_1)})_A,
\mathcal N_s(H_\theta^{(\ell_2)})_A
\right\}
\label{eq:source-condition}
\end{equation}
To make $T$ combine mid-level geometry with high-level semantics, let
$\mathcal I$ denote the selected target-encoder layers and $w_\ell$ their
fusion weights. Each layer is spatially normalized over $V$ to align feature
scales across depths; the layers are then fused, normalized again, and
restricted to the candidate positions $Q$:
\begin{equation}
T=\left.
\mathcal N_s\!\left(
\sum_{\ell\in\mathcal I}w_\ell
\mathcal N_s(H_{\bar\theta,V}^{(\ell)})
\right)\right|_Q,
\quad \sum_{\ell}w_\ell=1
\label{eq:multilevel-target}
\end{equation}
Here, $C$ conditions $E_\phi(C,q)$, while $T$ supervises
$\mathcal L_{\mathrm{field}}$ and provides the reference state in
$\mathcal L_{\mathrm{enc}}$.

\subsection{Learning a Reachable Latent Energy Field}

For the neural energy parameterization, a Conditional Energy Transformer parameterizes the
scalar energy $E_\phi(C,q)\in\mathbb R$. Self-attention models interactions
among candidate states in $Q$, while cross-attention conditions them on the
multi-level anchors in $C$. A token-wise linear head shared across positions
maps each contextualized candidate token $h_i$ to a local contribution $e_i$,
which are summed to obtain
\begin{equation}
E_\phi(C,q)=\sum_i e_i
\label{eq:neural-scalar-energy}
\end{equation}
To shape the local landscape around the data manifold, the clean candidate
representation $S$ is perturbed by scale-preserving Gaussian corruption:
\begin{equation}
q_0=\sqrt{1-\sigma^2}\,S+\sigma\varepsilon,\qquad
\varepsilon\sim\mathcal N(0,I)
\label{eq:corruption}
\end{equation}
where $\sigma$ denotes the corruption magnitude.
Starting from $q_0$, we perform $K$ clipped descent updates on the energy
surface,
\begin{equation}
g_k=\nabla_{q_k}E_\phi(C,q_k),\qquad
q_{k+1}=q_k-\alpha\,\mathcal C_\tau(g_k),
\label{eq:energy-gradient-updates}
\end{equation}
where $\mathcal C_\tau$ clips the gradient RMS of each sample to at most
$\tau$, limiting the RMS displacement of one update to $\alpha\tau$. Let
$\Delta=\operatorname{sg}(T-q_0)$ be the ideal recovery vector,
$r=\operatorname{RMS}(\Delta)$ its distance, and
$U=\Delta/(r+\epsilon)$ its RMS-normalized direction.

Requiring $q_K$ to reach $T$ within $K$ steps would demand a per-step RMS of
$r/K$. Under stronger corruption, $r/K$ may exceed the distance attainable by
a clipped update and would therefore distort the desired energy gradient. We
use a reachable-displacement schedule that restricts the supervised step size
to a fraction $\beta\in(0,1]$ of the maximum update:
\begin{equation}
\rho=\min\!\left(\frac{r}{K},\,\beta\alpha\tau\right),
\qquad D_k=k\rho U
\label{eq:reachable-schedule}
\end{equation}
If the required per-step displacement $r/K$ does not exceed the permitted
supervised step size $\beta\alpha\tau$, the cumulative supervised displacement
at step $K$ covers the full recovery path from $q_0$ to $T$; otherwise, each
step uses the maximum permitted magnitude and constructs the supervised
displacement along the direction from $q_0$ toward $T$. The Energy Field Loss matches the cumulative displacement induced by
the energy updates, $q_k-q_0$, to $D_k$:
\begin{equation}
\mathcal L_{\mathrm{field}}
=\frac{1}{\sum_{k=1}^{K}a_k}
\sum_{k=1}^{K}a_k
\frac{\|(q_k-q_0)-D_k\|_{Q}^{2}}{|Q|D}
\label{eq:field-loss}
\end{equation}
Supervising cumulative displacement rather than absolute state trains the
energy field to point toward $T$ at each step without requiring an intermediate
$q_k$ to equal the target. Both $S$ and $C$ are stop-gradient inputs in
$\mathcal L_{\mathrm{field}}$, so this path updates only the energy parameters
$\phi$. Because each updated state $q_k$ depends on $\nabla_q E$, optimizing
$\mathcal L_{\mathrm{field}}$ requires backpropagating through these
energy-gradient updates and therefore involves mixed second-order derivatives
with respect to the state and energy parameters; automatic differentiation
computes only the required Hessian--vector products without explicitly forming
the full Hessian.

\subsection{Relative-Energy Supervision for the Encoder}

The visual encoder must translate the learned structural preference into its
feed-forward representations so that its output $S$ already occupies a
coherent low-energy region.
Minimizing $E_\phi(C,S)$ in isolation provides no sample-specific reference and
admits trivial collapse of the energy scale. Because the multi-level target
$T$ represents an uncorrupted, structurally coherent configuration under the
same context $C$, LEF-JEPA evaluates $S$ and $T$ with the same energy function
and uses the target energy as a sample-specific comparison reference. Let
$\bar E(C,X)=E(C,X)/|Q|$ denote energy
normalized by the number of candidate positions. A smooth one-sided objective
defines the Encoder Energy Loss:
\begin{equation}
\mathcal L_{\mathrm{enc}}
=\gamma\,\operatorname{softplus}\!\left(
\frac{\bar E_\phi(\operatorname{sg}(C),S)-
\operatorname{sg}[\bar E_\phi(C,T)]}{\gamma}
\right)
\label{eq:encoder-energy}
\end{equation}
Here, $\gamma>0$ controls the smoothness of the penalty. The softplus places
the strongest pressure on positive relative-energy gaps and progressively
attenuates once $\bar E_\phi(C,S)$ falls below $\bar E_\phi(C,T)$. It neither
forces $S=T$ nor requires the two scalar energies to match; instead, it
constrains $S$ to the low-energy side defined by the target reference. In
$\mathcal L_{\mathrm{enc}}$, the energy parameters, the explicit anchor
condition $C$, and the target representation $T$ are stop-gradient. The loss
updates the visual encoder through $S$, without differentiating through the
$K$-step recovery process. Since $S$ is normalized using statistics shared
across all visible tokens, the resulting encoder gradient remains spatially
coupled across visible positions.

The parameter-specific objectives are
\begin{equation}
\min_{\theta,\psi}\;
\mathcal L_{\mathrm{pred}}+\lambda_{\mathrm{enc}}
\mathcal L_{\mathrm{enc}},
\qquad
\min_{\phi}\;\lambda_{\mathrm{field}}
\mathcal L_{\mathrm{field}}
\label{eq:separated-objectives}
\end{equation}
where $\lambda_{\mathrm{enc}}$ balances the encoder energy constraint with the
masked-prediction objective; after pretraining, the auxiliary energy module is
discarded entirely.

\begin{figure*}[t]
  \centering
  \includegraphics[width=0.88\textwidth]{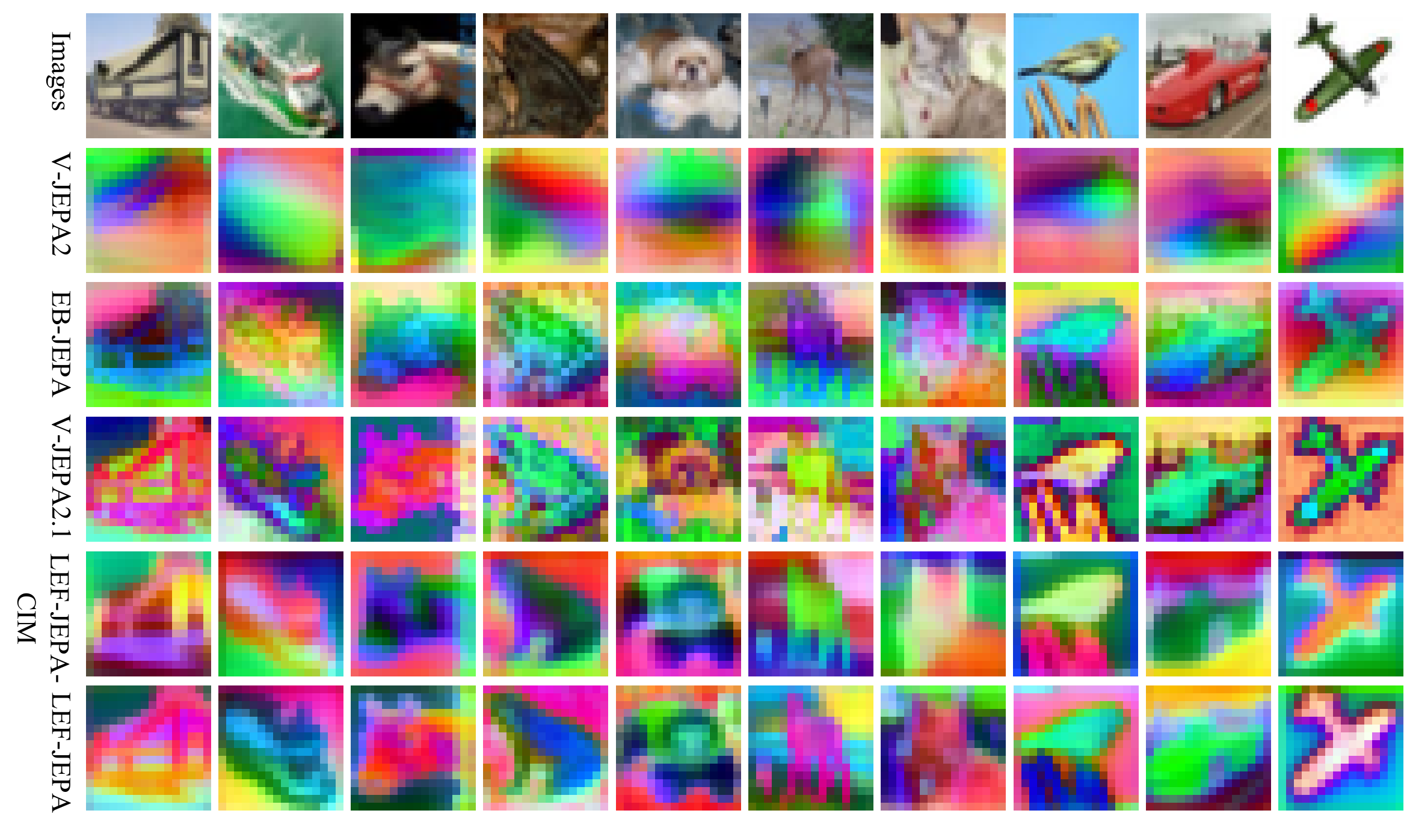}
  \caption{Patch-token PCA on CIFAR-10 samples. LEF-JEPA produces
  coherent regions aligned with object extent and boundaries. PCA is fitted
  independently per image, so colors are not comparable across methods.}
  \label{fig:pca-comparison}
\end{figure*}

\subsection{Conditional Ising Energy for Specialized Ising Solvers}
\label{sec:ising-realization}

This Hamiltonian form is natively compatible with specialized Ising solvers,
including superconducting quantum annealers and coherent Ising machines. In
this work, we integrate a physical coherent Ising machine (CIM) into the closed
training loop to solve hardware-compatible Hamiltonians and return a pool
$\mathcal S$ of low-energy states. A coefficient network maps $(C,q)$ to unary
biases $h_i(C,q)$ and symmetric
pairwise couplings $J_{ij}(C,q)=J_{ji}(C,q)$ with zero diagonal, defining a
Hamiltonian over $B=16$ logical binary spins:
\begin{equation}
\mathcal H_\omega(s;C,q) = -\sum_{i=1}^{B}h_i s_i
- \frac{1}{2}\sum_{i\ne j}J_{ij}s_is_j,
\quad s_i\in\{-1,+1\}
\label{eq:ising-hamiltonian}
\end{equation}
Each forward evaluation connects discrete physical search with continuous
backpropagation in four stages: (i) a classical network predicts continuous
coefficients $(h,J)$; (ii) the photonic CIM searches the quantized Hamiltonian
and returns a compact pool $\mathcal S$ of low-energy spin configurations;
(iii) the discrete states are rescored by the original, differentiable
coefficients; and (iv) a truncated soft minimum aggregates the pool into a
smooth conditional scalar energy,
\begin{equation}
E_\omega^{\mathrm{Ising}}(C,q)
=-\tau_s\log\!\left[
\frac{1}{|\mathcal S|}
\sum_{s\in\mathcal S}
\exp\!\left[-\mathcal H_\omega(s;C,q)/\tau_s\right]
\right]
\label{eq:ising-softmin}
\end{equation}
During differentiation, each $s\in\mathcal S$ is held fixed as a structural
anchor, while gradients propagate through $(h,J)$ to the latent representation
$q$. Aggregating the low-energy pool in Eq.~\eqref{eq:ising-softmin} smooths
changes between competing low-energy configurations.

Within a trajectory of $K$ energy updates, the physical CIM is queried only at
the initial perturbed state $q_0$; the returned pool $\mathcal S$ is reused for
all $K$ local updates. The number of physical calls is therefore
$\mathcal O(1)$ with respect to $K$. For $\mathcal L_{\mathrm{enc}}$, separate
low-energy pools are queried for $S$ and $T$, and their energies are compared
through Eq.~\eqref{eq:encoder-energy}. The photonic solver thus supplies global
discrete low-energy candidates that enter self-supervised representation
learning through differentiable rescoring.

 \section{Experiments}
\label{sec:experiments}

\subsection{Experimental Settings}
\label{sec:experimental-settings}

\paragraph{Unified pretraining setup and energy realizations.}
To compare architectures under matched conditions, all methods pretrain their
visual encoders from scratch with self-supervision on unlabeled CIFAR-10
~\cite{krizhevsky2009cifar}. Despite the $32\times32$ input resolution, we use
$2\times2$ patches, producing a dense $16\times16$ spatial token grid (256
tokens per image) that provides fine-grained spatial sampling for modeling
object contours and local structure. We evaluate two realizations of our
framework: LEF-JEPA with neural scalar energy, and LEF-JEPA-CIM with a
conditional Ising Hamiltonian solved by a physical photonic CIM. For each
energy realization, $K$ and $\lambda_{\mathrm{enc}}$ are selected according to
linear-classification accuracy on a fixed 5,000-image validation subset of the
CIFAR-10 training split and then fixed for all subsequent downstream
evaluations. Hardware specifications, quantization protocol, solver timing,
and hyperparameter-selection details are provided in the supplementary
material.

\noindent\textbf{Frozen-encoder evaluation at three task granularities.}
After pretraining, each visual encoder is frozen and evaluated at three levels.
CIFAR-10 linear classification measures in-distribution global semantics and
reports top-1 accuracy. Out-of-distribution dense transfer to PartImageNet
~\cite{he2022partimagenet} trains a linear foreground-segmentation head on
frozen patch tokens and reports foreground IoU. Action-conditioned spatial
planning is evaluated through goal-reaching trajectories in the continuous
Two-Room environment.

\subsection{Joint Semantic and Spatial Gains}
\label{sec:main-results}

\begin{table*}[t]
  \centering
  \caption{Frozen-encoder transfer across classification, foreground
  segmentation, and visual planning. Results aggregate three independently
  pretrained encoders; classification accuracy and foreground IoU are reported
  as mean $\pm$ standard deviation across encoders. Each encoder is evaluated
  on 100 Two-Room episodes, and SR is the mean success rate across the three
  experiments, rounded to the nearest percentage point. Best and second-best
  within each backbone are bold and underlined.}
  \label{tab:main-downstream}
  \small
  \setlength{\tabcolsep}{4pt}
  \begin{tabular*}{\textwidth}{@{\extracolsep{\fill}}lcccccc@{}}
    \toprule
    \multirow{2}{*}{Method}
    & \multicolumn{3}{c}{ViT-S} & \multicolumn{3}{c}{ViT-B} \\
    \cmidrule(lr){2-4}\cmidrule(lr){5-7}
    & Acc. (\%) $\uparrow$ & Fg. IoU (\%) $\uparrow$ & SR (\%) $\uparrow$
           & Acc. (\%) $\uparrow$ & Fg. IoU (\%) $\uparrow$ & SR (\%) $\uparrow$ \\
    \midrule
    V-JEPA2~\cite{assran2025vjepa2} (no LEF)
      & $89.40 \pm 0.025$ & $53.54 \pm 0.037$ & $65$
      & $89.51 \pm 0.024$ & $64.29 \pm 0.038$ & $73$ \\
    EB-JEPA~\cite{terver2026ebjepa}
      & $81.49 \pm 0.040$ & $60.29 \pm 0.089$ & $59$
      & $85.89 \pm 0.017$ & $71.15 \pm 0.007$ & $72$ \\
    V-JEPA2.1~\cite{murlabadia2026vjepa21}
      & $83.99 \pm 0.024$ & $65.30 \pm 0.012$ & $53$
      & $88.77 \pm 0.005$ & $67.47 \pm 0.045$ & $77$ \\
    \midrule
    LEF-JEPA-CIM
      & $\boldsymbol{91.80 \pm 0.005}$ & \underline{$68.44 \pm 0.039$} & $\boldsymbol{79}$
      & $\boldsymbol{92.27 \pm 0.010}$ & $\boldsymbol{72.11 \pm 0.044}$ & $\boldsymbol{86}$ \\
    LEF-JEPA
      & \underline{$91.63 \pm 0.016$} & $\boldsymbol{70.04 \pm 0.040}$ & \underline{$71$}
      & \underline{$92.17 \pm 0.009$} & \underline{$71.44 \pm 0.021$} & \underline{$84$} \\
    \bottomrule
  \end{tabular*}
\end{table*}

\begin{figure}[b]
  \centering
  \includegraphics[width=\linewidth]{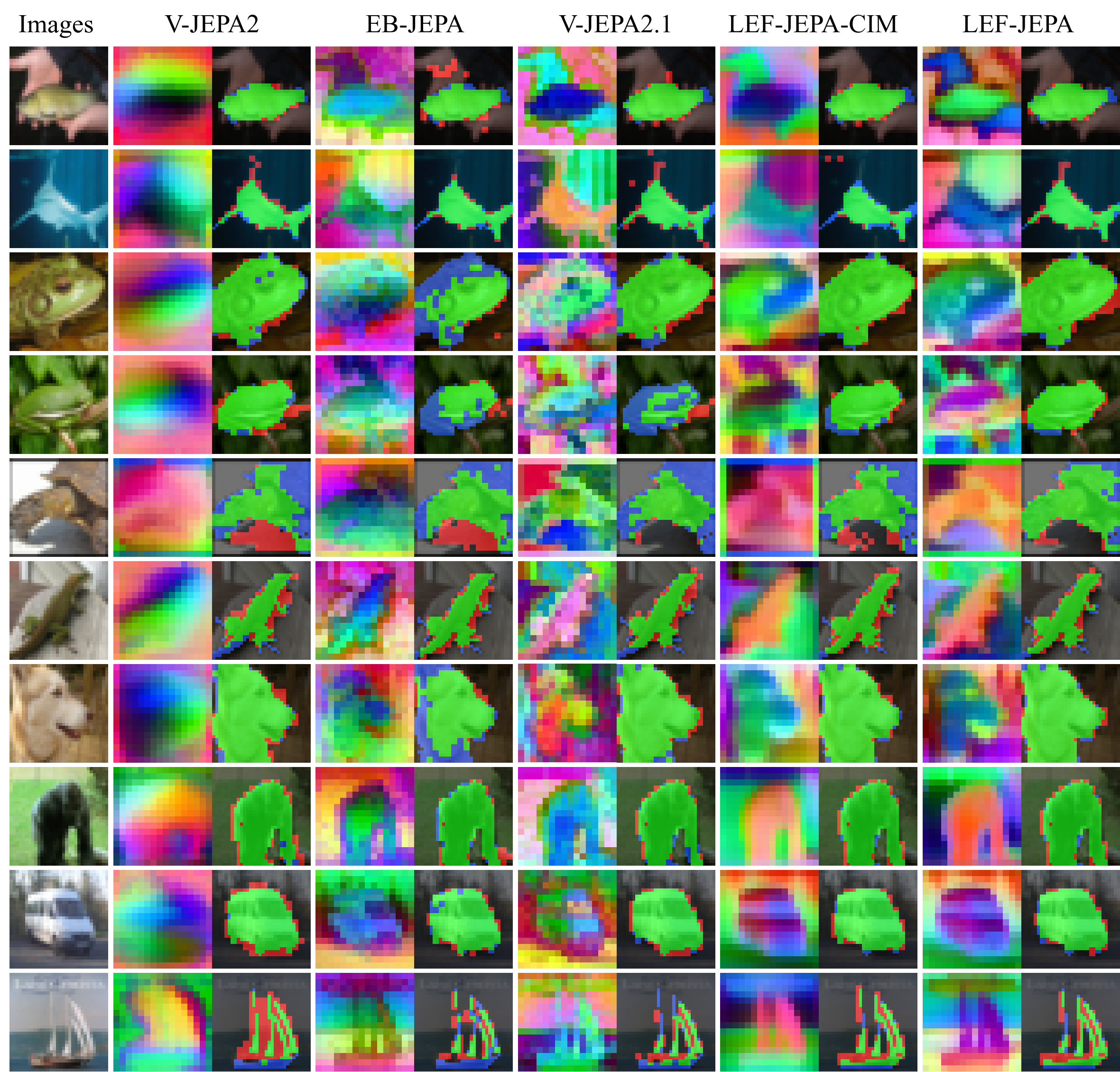}
  \caption{Spatial representations and foreground segmentation on PartImageNet.
  Each method is shown with patch-token PCA and foreground predictions. Green,
  red, and blue denote
  true positives, false positives, and false negatives. LEF-JEPA yields more
  complete masks and sharper boundaries.}
  \label{fig:segmentation-comparison}
\end{figure}

\paragraph{Latent organization and cross-task generalization.}
Table~\ref{tab:main-downstream} shows that on ViT-S, V-JEPA2.1 improves
foreground IoU over V-JEPA2 by $11.76$ points but reduces classification
accuracy and planning success by $5.41$ and $12$ points; EB-JEPA likewise
improves segmentation while lowering classification. With frozen encoders and
matched task heads, both LEF realizations outperform these baselines across
both backbones and all three tasks. On ViT-B, for example, LEF-JEPA improves
classification accuracy, foreground IoU, and planning success over V-JEPA2.1
by $3.40$, $3.97$, and $7$ points, respectively.

\begin{figure}[H]
  \centering
  \includegraphics[width=\linewidth]{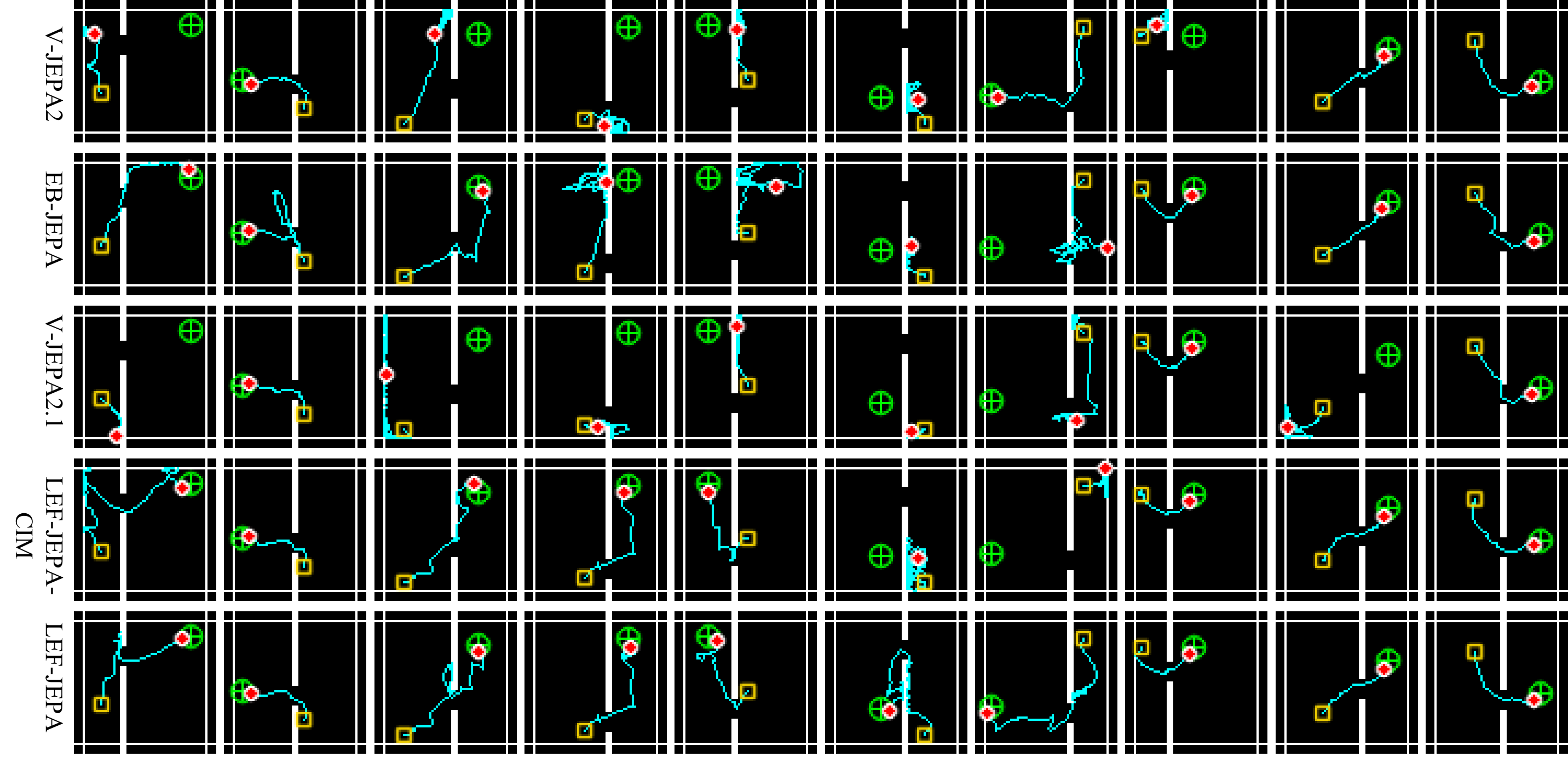}
  \caption{Two-Room planning trajectories. Each column fixes the same task
  across methods. Yellow, green, cyan,
  and red denote the start, goal, executed path, and final state.}
  \label{fig:two-room-comparison}
\end{figure}

\paragraph{Spatial representations and task visualizations.}
Patch-token PCA in Figure~\ref{fig:pca-comparison} shows that LEF-JEPA preserves
clearer contours and internal structure than V-JEPA2; compared with V-JEPA2.1,
it responds less to local texture and color variations and forms more coherent
object regions. The out-of-distribution PartImageNet samples exhibit more complete
foreground masks, with fewer spills and holes around narrow structures
(Figure~\ref{fig:segmentation-comparison}); in the shown Two-Room trajectories,
both LEF realizations pass through the doorway and reach the goal region
(Figure~\ref{fig:two-room-comparison}).

\noindent\textbf{Neural-energy and physical-Ising realizations.}
As shown in Table~\ref{tab:main-downstream}, the two realizations shape
representation geometry in fundamentally different ways through a continuous
gradient field and a discrete-spin Hamiltonian. In continuous latent space,
LEF-JEPA accumulates local recovery displacements through iterative descent
($K=3$). This fine-grained continuous guidance is particularly important for
dense boundary parsing with the smaller ViT-S, yielding the highest foreground
IoU of $70.04\%$. In contrast, LEF-JEPA-CIM casts local correction as a
combinatorial low-energy search on a physical CIM: discrete spin couplings
naturally filter fine-grained local texture and color fluctuations, directly
providing a region-level consistency constraint. Consequently, with a single
local update ($K=1$), LEF-JEPA-CIM efficiently guides the encoder, attaining the
strongest global classification and spatial planning results on both backbones.
On ViT-B, LEF-JEPA-CIM further outperforms LEF-JEPA and achieves the highest
foreground IoU of $72.11\%$.

\subsection{Ablation Studies}
\label{sec:ablation-studies}

\subsubsection{Number of Energy-Gradient Updates}
\label{sec:ablation-update-count}

\begin{figure}[t]
  \centering
  \includegraphics[width=\linewidth]{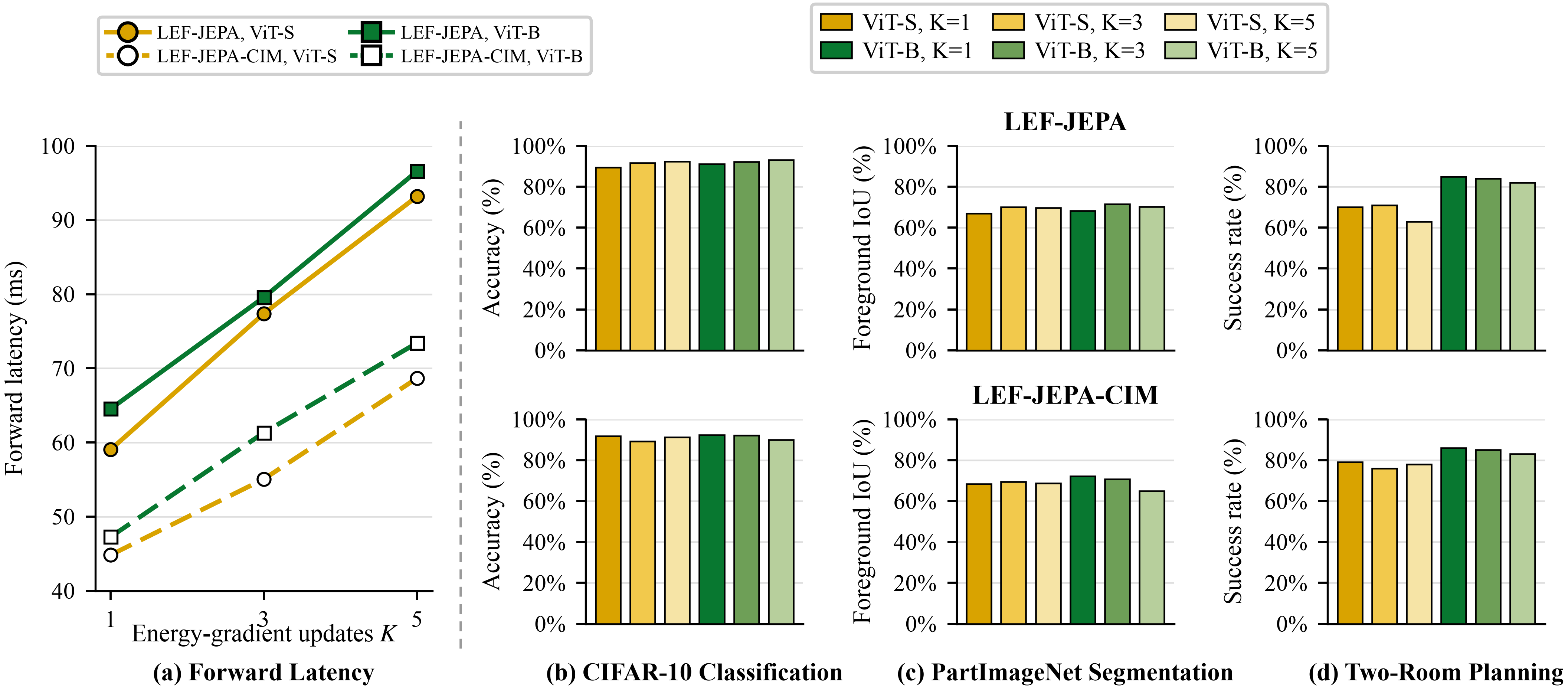}
  \caption{Effect of the number of energy-gradient updates $K$ on pretraining
  latency and downstream transfer. Each CIM trajectory uses one physical solve
  and reuses its low-energy state pool across all $K$ updates. CIM latency
  excludes cloud communication and queueing; bars report means over three
  independently pretrained encoders.}
  \label{fig:k-step-combined}
\end{figure}

Figure~\ref{fig:k-step-combined} compares the effect of the number of
energy-gradient updates $K\in\{1,3,5\}$ on pretraining latency and downstream
transfer. Even with a single update ($K=1$), both energy realizations improve
segmentation and planning over the energy-free V-JEPA2 baseline. For continuous
neural energy, $K=3$ attains the highest foreground IoU on both backbones and
provides a more balanced cross-task profile; increasing to $K=5$ further
improves classification accuracy but does not further improve segmentation or
planning performance.

In contrast, LEF-JEPA-CIM favors a shallower update depth: $K=1$ gives the best
results on all three tasks for ViT-B, while classification and planning are
likewise best at $K=1$ for ViT-S. Because the physical CIM is queried only once
at the initial state $q_0$ and subsequent updates reuse the same low-energy spin
pool, increasing $K$ primarily repeats rescoring and yields no consistent gain.

\subsubsection{Weight of the Relative-Energy Constraint
\texorpdfstring{$\lambda_{\mathrm{enc}}$}{lambda enc}}
\label{sec:ablation-encoder-energy}

Table~\ref{tab:energy-scale-ablation} compares the effect of different
$\lambda_{\mathrm{enc}}$ values on downstream transfer with ViT-S, where
LEF-JEPA and LEF-JEPA-CIM use $K=3$ and $K=1$, respectively. This coefficient
controls the strength of the encoder energy constraint
$\mathcal L_{\mathrm{enc}}$ relative to the masked-prediction objective
$\mathcal L_{\mathrm{pred}}$.

\begin{table}[t]
  \centering
  \caption{Encoder-energy coefficient ablation with ViT-S. LEF-JEPA uses
  $K=3$, whereas LEF-JEPA-CIM uses $K=1$.
  Classification and foreground IoU are mean $\pm$ standard deviation across
  three independently pretrained encoders; each encoder is evaluated on 100
  Two-Room episodes, and SR is averaged across the three runs and rounded to
  the nearest percentage point. Bold marks the best result within each
  realization.}
  \label{tab:energy-scale-ablation}
  \tiny
  \setlength{\tabcolsep}{0.15pt}
  \begin{tabular*}{\columnwidth}{@{\extracolsep{\fill}}cccccccc@{}}
    \toprule
    \multicolumn{4}{c}{LEF-JEPA} & \multicolumn{4}{c}{LEF-JEPA-CIM} \\
    \cmidrule(lr){1-4}\cmidrule(lr){5-8}
    $\lambda_{\mathrm{enc}}$ & Acc. (\%) $\uparrow$
    & Fg. IoU (\%) $\uparrow$ & SR (\%) $\uparrow$
    & $\lambda_{\mathrm{enc}}$ & Acc. (\%) $\uparrow$
    & Fg. IoU (\%) $\uparrow$ & SR (\%) $\uparrow$ \\
    \midrule
    $0.001$ & $\mathbf{92.40}\!\pm\!\mathbf{.010}$ & $69.20\!\pm\!.014$ & $69$
    & $5\!\times\!10^{-5}$ & $89.09\!\pm\!.026$ & $66.67\!\pm\!.024$ & $79$ \\
    $0.005$ & $91.63\!\pm\!.016$ & $\mathbf{70.04}\!\pm\!\mathbf{.040}$ & $71$
    & $10^{-4}$ & $\mathbf{91.80}\!\pm\!\mathbf{.005}$ & $\mathbf{68.44}\!\pm\!\mathbf{.039}$ & $79$ \\
    $0.009$ & $84.04\!\pm\!.050$ & $61.11\!\pm\!.060$ & $\mathbf{80}$
    & $0.005$ & $71.82\!\pm\!.054$ & $47.19\!\pm\!.083$ & $53$ \\
    $0.05$ & $39.38\!\pm\!.070$ & $31.66\!\pm\!.014$ & $0$
    & $0.05$ & $68.25\!\pm\!.032$ & $48.83\!\pm\!.013$ & $\mathbf{81}$ \\
    \bottomrule
  \end{tabular*}
\end{table}

As shown in Table~\ref{tab:energy-scale-ablation}, the task-specific optima do
not coincide for continuous energy: $\lambda_{\mathrm{enc}}=0.005$ attains the
highest foreground IoU while maintaining strong overall performance in
classification and planning. For LEF-JEPA-CIM, $10^{-4}$ yields the highest
classification accuracy and foreground IoU while retaining $79\%$ planning
success. Increasing the coefficient to $0.05$ raises planning success to $81\%$
but substantially reduces classification and segmentation. Overall, both
realizations show clear sensitivity to the constraint strength, and an overly
strong relative-energy constraint degrades classification and segmentation
performance.

 \section{Conclusion}
\label{sec:conclusion}

This work shows that latent representations for world modeling should not only
recover the spatial structure of objects, but also avoid overemphasizing
low-level visual information, such as texture and color, that is not always
necessary for prediction and action. Through
target-guided latent recovery and relative-energy supervision, LEF-JEPA enables
frozen representations to benefit jointly across image recognition,
cross-domain foreground segmentation, and visual planning without adding
downstream inference overhead. Our experiments further show that this energy
learning framework can be realized either by a continuous neural network or by
coupling a conditional Ising Hamiltonian to a physical CIM; the latter retains
strong cross-task performance with a shallow local-update horizon. Overall,
the results show that structure learning in latent space need not rely on the
pointwise modeling of low-level visual information such as texture and color;
instead, local recovery and energy relations can organize the spatial structure
required by downstream tasks. Future work will explore the applicability and
efficiency of this mechanism in long-video world models and on larger-scale
specialized Ising solvers.

 {
    \small
    \bibliographystyle{ieeenat_fullname}
    \bibliography{main}
}

\clearpage
\onecolumn
\setcounter{figure}{0}
\setcounter{table}{0}
\setcounter{equation}{0}
\renewcommand{\thefigure}{S\arabic{figure}}
\renewcommand{\thetable}{S\arabic{table}}
\renewcommand{\theequation}{S\arabic{equation}}
\renewcommand{\theHfigure}{supp.figure.\arabic{figure}}
\renewcommand{\theHtable}{supp.table.\arabic{table}}
\renewcommand{\theHequation}{supp.equation.\arabic{equation}}
\appendix
\section*{Supplementary Material}

\section{Notation and Optimization Paths}
\label{supp:notation}

\subsection{Symbols}

Table~\ref{tab:symbols} collects the main notation used across the neural and
Ising energy realizations. The symbol $S$ denotes the clean online candidate
representation, whereas $\mathcal S$ denotes a finite pool of Ising spin
states; the different typefaces are intentional.

\begin{table}[H]
  \centering
  \caption{Notation used in the main paper and supplement.}
  \label{tab:symbols}
  \small
  \setlength{\tabcolsep}{5pt}
  \begin{tabular*}{\textwidth}{@{\extracolsep{\fill}}p{0.13\textwidth}p{0.34\textwidth}p{0.13\textwidth}p{0.34\textwidth}@{}}
    \toprule
    Symbol & Meaning & Symbol & Meaning \\
    \midrule
    $x$ & Input image & $\Omega$ & Set of all patch positions \\
    $V,M$ & Visible and masked patch positions & $Q,A$ & Disjoint candidate and anchor subsets of $V$ \\
    $f_\theta$ & Online visual encoder & $f_{\bar\theta}$ & EMA-updated target encoder \\
    $p_\psi$ & Masked latent predictor & $H_\theta^{(\ell)}$ & Online patch tokens at encoder layer $\ell$ \\
    $\mathcal N_s$ & Spatial centering followed by one image-level RMS scaling & $S$ & Clean online candidate representation \\
    $C$ & Multi-level online anchor condition & $T$ & Multi-level target representation at candidate positions \\
    $E_\phi(C,q)$ & Neural conditional scalar energy & $q_0$ & Scale-preserving Gaussian perturbation of $S$ \\
    $q_k$ & Candidate state after $k$ energy-gradient updates & $K$ & Number of energy-gradient updates \\
    $\alpha$ & Update step size & $\tau$ & Per-sample RMS threshold for clipping $\nabla_qE$ \\
    $\sigma$ & Corruption magnitude & $D_k$ & Reachable target displacement at update $k$ \\
    $\mathcal C_\tau$ & Per-sample RMS gradient-clipping operator & $\beta$ & Reachable fraction of the maximum clipped update \\
    $\rho$ & Reachable per-step RMS displacement & $\gamma$ & Smoothness scale of the one-sided Encoder Energy Loss \\
    $\bar E(C,X)$ & Energy normalized by $|Q|$ & $\phi,\omega$ & Neural and Ising energy parameters \\
    $a_k$ & Weight of update $k$ in the Energy Field Loss & $\lambda_{\mathrm{enc}}$ & Encoder Energy Loss weight \\
    $\mathcal L_{\mathrm{pred}}$ & Masked Prediction Loss & $\mathcal L_{\mathrm{field}}$ & Energy Field Loss \\
    $\mathcal L_{\mathrm{enc}}$ & Encoder Energy Loss & $\lambda_{\mathrm{field}}$ & Energy Field Loss weight \\
    $\mathcal H_\omega(s;C,q)$ & Conditional Ising Hamiltonian & $E_\omega^{\mathrm{Ising}}(C,q)$ & Soft-min scalar energy over the state pool \\
    $h,J$ & Unary biases and pairwise Ising couplings & $s\in\{-1,+1\}^{B}$ & A logical Ising state with $B=16$ spins \\
    $\mathcal S$ & Solver-proposed low-energy spin-state pool & $\tau_s$ & Temperature of the truncated soft minimum \\
    \bottomrule
  \end{tabular*}
\end{table}

\subsection{Gradient routing}

The two energy objectives share the same conditional energy function but
assign it different optimization roles. Table~\ref{tab:gradient-routing}
lists the trainable recipients of each loss. The Energy Field Loss treats
$S$, $C$, and $T$ as fixed and updates only the energy parameters. Its
$K$-step update graph retains $\nabla_q E$, so backpropagation evaluates the
required Hessian--vector products involving mixed second-order derivatives with
respect to the state and energy parameters, without materializing a full
Hessian. The Encoder
Energy Loss freezes the energy parameters, the explicit anchor-condition
branch, and the target representation. Gradients enter the online encoder
through the source branch $S$; the shared visible-token centering and RMS
statistics used to construct $S$ couple candidate and anchor positions. This
separation prevents the unrolled energy-gradient updates from becoming an
additional reconstruction path into the visual encoder.

\begin{table}[H]
  \centering
  \caption{Gradient recipients for the three training objectives. A check mark
  indicates that the corresponding parameters are updated by that loss.}
  \label{tab:gradient-routing}
  \small
  \setlength{\tabcolsep}{4pt}
  \begin{tabular}{lccc}
    \toprule
    Objective & Online encoder $\theta$ & Predictor $\psi$ & Energy $\phi/\omega$ \\
    \midrule
    $\mathcal L_{\mathrm{pred}}$  & \checkmark & \checkmark & -- \\
    $\mathcal L_{\mathrm{field}}$ & -- & -- & \checkmark \\
    $\mathcal L_{\mathrm{enc}}$   & \checkmark & -- & -- \\
    \bottomrule
  \end{tabular}
\end{table}

\subsection{Local-roughness visualization}

The local-roughness maps in Figure~1 are computed from final-layer patch
tokens without training a probe. For each image, we fit PCA independently and
retain the first three components. Let $z_p\in\mathbb R^3$ be the projected
coordinate at patch position $p$, let $\tilde z_p$ denote its spatially centered
value, and define the image-level RMS-normalized coordinate
\begin{equation}
u_p=\frac{\tilde z_p}{
\sqrt{|\Omega|^{-1}\sum_{r\in\Omega}\|\tilde z_r\|_2^2}}.
\label{eq:supp-pca-rms}
\end{equation}
The local roughness at $p$ is the mean squared difference to its available
horizontal and vertical one-patch neighbors:
\begin{equation}
R(p)=\frac{1}{|\mathcal N_4(p)|}
\sum_{q\in\mathcal N_4(p)}\|u_p-u_q\|_2^2,
\label{eq:supp-local-roughness}
\end{equation}
where $\mathcal N_4(p)$ contains the valid four-neighborhood positions. Thus,
bright responses indicate a stronger local change in the three-dimensional PCA
field. All methods use the same fixed visualization range $R\in[0,2]$.

\subsection{Pretraining hyperparameters}

Table~\ref{tab:pretraining-hparams} summarizes the hyperparameters used in the
main pretraining experiments. Batch sizes are per device. The online optimizer
jointly updates the visual encoder and masked predictor, while a separate
optimizer updates the energy module.

\begin{table}[H]
  \centering
  \caption{Pretraining hyperparameters for the two energy realizations.}
  \label{tab:pretraining-hparams}
  \small
  \setlength{\tabcolsep}{3.5pt}
  \begin{tabular}{lcc}
    \toprule
    Hyperparameter & LEF-JEPA & LEF-JEPA-CIM \\
    \midrule
    Corruption magnitude $\sigma$ & $\mathcal U(0.1,0.5)$ & $\mathcal U(0.1,0.5)$ \\
    Update step size $\alpha$ & $0.1$ & $0.1$ \\
    RMS clipping threshold $\tau$ & $1.0$ & $1.0$ \\
    Reachable fraction $\beta$ & $0.8$ & $0.8$ \\
    Pretraining epochs & $300$ & $300$ \\
    Batch size (ViT-S / ViT-B) & $256/256$ & $256/256$ \\
    Optimizer & AdamW (both) & AdamW (both) \\
    Learning rate & $3\!\times\!10^{-4}$ (both) & $3\!\times\!10^{-4}$ (both) \\
    \bottomrule
  \end{tabular}
\end{table}

\section{Datasets and Task Construction}
\label{supp:datasets}

\subsection{CIFAR-10 pretraining and classification}

CIFAR-10 contains $50{,}000$ training images and $10{,}000$ test images from
ten object categories, each at $32\times32$ resolution. Encoder parameters are
optimized exclusively by self-supervised objectives over the complete CIFAR-10
training split. For hyperparameter selection, we use a fixed 5,000-image subset
of the training split as the validation subset and fit a linear classifier on
frozen representations from the remaining training images. We select $K$ and
$\lambda_{\mathrm{enc}}$ using the mean top-1 accuracy of three independently
pretrained encoders on the validation subset. The selected hyperparameters are
then fixed for all subsequent downstream evaluations. The online and target
encoders receive the same augmented view. Augmentation
comprises reflection-padded random cropping, horizontal flipping, and mild
color jitter, applied with probabilities $0.5$, $0.5$, and $0.2$,
respectively. Within each encoder scale, all methods use the same augmentation,
masking distribution, and number of pretraining epochs.

\subsection{PartImageNet foreground transfer}

PartImageNet provides dense part annotations for general object categories.
The evaluation converts its whole-object annotation
into a binary foreground/background target. The local release contains
$20{,}466$ training, $1{,}206$ validation, and $2{,}408$ test images. Images
and masks are resized to $32\times32$ using bilinear and nearest-neighbor
interpolation, respectively. A patch is labeled foreground when at least half
of its pixels belong to the annotated object. This conversion gives a
$16\times16$ target grid for a patch size of two pixels. The encoder remains
frozen and only a shared $1\times1$ two-class convolution is fitted over its
final patch features.

\subsection{Two-Room visual planning}

Two-Room is a controlled $65\times65$ environment containing an agent and a
vertical wall with a traversable opening. Observations contain separate agent
and wall channels. The image adapter renders these channels as RGB, resizes the
result to $32\times32$.

Each predictor step consumes a block of five consecutive two-dimensional
primitive actions. Training unrolls eight such blocks, corresponding to $40$
primitive actions and nine latent states. Evaluation is open loop. Start and
goal states are sampled as random pairs,
and success is declared when the Euclidean state distance falls below $4.5$
environment units.

\section{Architecture and Pretraining Details}
\label{supp:architecture}

The baselines are V-JEPA2 as the direct no-energy control, EB-JEPA retrained
through its official repository, and V-JEPA2.1 retrained from its open-source
implementation under the same settings.

\subsection{Network architecture and target representations}

\begin{table}[H]
  \centering
  \caption{Network architecture. The encoder, predictor, and energy network
  apply two-dimensional rotary position encoding to attention queries and
  keys.}
  \label{tab:architecture}
  \scriptsize
  \setlength{\tabcolsep}{3pt}
  \begin{tabular*}{\columnwidth}{@{\extracolsep{\fill}}lccc@{}}
    \toprule
    Component & Width & Blocks / heads & MLP \\
    \midrule
    ViT-S encoder & 384 & 12 / 6 & 4 \\
    ViT-B encoder & 768 & 12 / 12 & 4 \\
    Masked predictor & 160 & 2 / 8 & 2 \\
    Conditional energy Transformer & 160 & 2 / 4 & 4 \\
    \bottomrule
  \end{tabular*}
\end{table}

The masked region is the union of eleven or twelve overlapping
$4\times4$ blocks, calibrated to an average masking ratio of approximately
$50\%$. Visible positions are partitioned into candidate and anchor subsets at
a $1{:}1$ ratio using block-based spatial sampling, while retaining their
original grid coordinates.

The target encoder is initialized from the online encoder and updated by EMA.
Its momentum follows a cosine schedule from $0.996$ to $0.9995$, and target
tokens are normalized without learnable affine parameters. The anchor
condition uses online layers 6 and 9 with weights $(0.5,0.5)$; the target
reference combines target layers 6, 9, and 12 with weights $(0.25,0.25,0.5)$.
Each layer is spatially centered and RMS-normalized before fusion, and the
fused target representation is normalized once more.

Candidate features and each anchor level are projected from the encoder width
to 160 dimensions. Each energy block contains candidate self-attention,
cross-attention from candidate queries to the anchor keys and values, and an
MLP. The two anchor levels extract conditional information separately before
their outputs are fused. For LEF-JEPA, a shared scalar head is
applied to the final normalized candidate features, and its position-wise
contributions are summed to produce the conditional energy.

\section{Conditional Ising and Physical CIM Implementation}
\label{supp:cim}

\subsection{Coefficient network and differentiable rescoring}

The Ising realization first uses $B=16$ learned queries to extract descriptors
$r_i\in\mathbb R^{160}$ for 16 logical spins from the candidate tokens. Each
descriptor predicts a unary spin bias $h_i$ and is paired with the other spin
descriptors to predict couplings $J_{ij}$, yielding an Ising Hamiltonian
conditioned on the current visual representation.

Concretely, each $r_i$ is projected to a 64-dimensional vector
$v_i=W_pr_i$. For every spin pair $(i,j)$, $v_i$ and $v_j$ form the symmetric
descriptor
\begin{equation}
u_{ij}=\left[v_i+v_j,\;|v_i-v_j|,\;v_i\odot v_j\right]
\label{eq:supp-pair-descriptor}
\end{equation}
A shared Pair MLP maps $u_{ij}$ to the coupling $J_{ij}$ using LayerNorm,
$192\rightarrow64$, GELU, and $64\rightarrow1$. Because $u_{ij}=u_{ji}$, its
output satisfies $J_{ij}=J_{ji}$; the diagonal is set to zero.

Hardware-returned spin candidates are detached, rescored with the original
floating-point Hamiltonian coefficients, and aggregated into a conditional
energy using the truncated soft minimum defined in the main paper. Gradients
therefore pass through the coefficient network to the representation, but not
through the discrete search process.

\subsection{Hardware queries and timing}

Physical runs access QBoson's CIM through the Kaiwu SDK. Each 16-spin logical
Hamiltonian is augmented with one reference spin, scaled by a detached positive
factor, and quantized to signed integers in $[-127,127]$. Up to 58 such
17-spin blocks are packed into one block-diagonal task, keeping each submission
within the 1000-spin hardware limit. Sampling mode returns ten complete
readouts of the packed system and therefore ten candidates for every 16-spin
Ising subproblem. After reference-spin decoding, candidates are ranked under the
submitted quantized Hamiltonian and deduplicated; up to eight distinct
low-energy states form the candidate pool.

Each local recovery trajectory queries its pool at the initial state $q_0$ and
reuses it across all $K$ updates; the coefficients are still recomputed and
rescored at every current state $q_k$.

All reported LEF-JEPA-CIM results use the physical solver. Pretraining forward
timing includes energy-network computation, coefficient preparation, physical
solving, and local Hamiltonian and state processing, but excludes network
transmission, service queueing, and result-return latency.

\section{Frozen Evaluation Protocols}
\label{supp:evaluation}

Within each backbone size, all methods use identically structured task heads
and matched optimization protocols; only the pretrained weights of the frozen
encoder differ.

All reported means and standard deviations aggregate three independently
pretrained encoders. Each pretraining seed controls the initialization of the
encoder and energy module, minibatch order, data augmentation, masking, and
candidate--anchor sampling. Downstream heads use a fixed initialization and
data order with the same optimization protocol for every encoder; the reported
seed variation therefore reflects encoder pretraining rather than downstream
head randomness.

\subsection{Classification and foreground segmentation}

All downstream evaluations freeze the EMA target encoder. For CIFAR-10
classification, final-layer patch tokens are spatially averaged, and the
resulting image-level features are standardized with training-set statistics
before fitting a linear classifier. The classifier is trained for 300 epochs
with AdamW and a cosine learning-rate schedule; accuracy is aggregated over
the three independently pretrained encoders.

For PartImageNet, final-layer patch tokens are cached and used to train a
shared linear $1\times1$ foreground head for 100 epochs. Test-set foreground
IoU is aggregated over the three independently pretrained encoders. Dataset splits and
foreground-label construction are described in Section~\ref{supp:datasets}.

\subsection{Two-Room visual planning}

Every method uses the same trajectories to train an identically structured
action-conditioned latent predictor. The predictor has two Transformer blocks
and eight attention heads, with the same width as the frozen encoder. Its
training objective combines one-step teacher forcing with an eight-block
recursive rollout.

Planning uses a horizon of eight latent blocks, corresponding to 40 primitive
actions. CEM performs 30 iterations over 300 candidates and retains 30 elites.
Candidate plans are ranked by the mean-squared error between terminal patch
tokens and the target representation. Evaluation is open loop.
Each pretrained encoder is evaluated on 100 episodes under the same
environment-generation protocol, and success rate is averaged across the three
encoder runs.

\section{Number of Energy-Gradient Updates}
\label{supp:energy-gradient-updates}

\begin{figure}[H]
  \centering
  \includegraphics[width=0.92\textwidth]{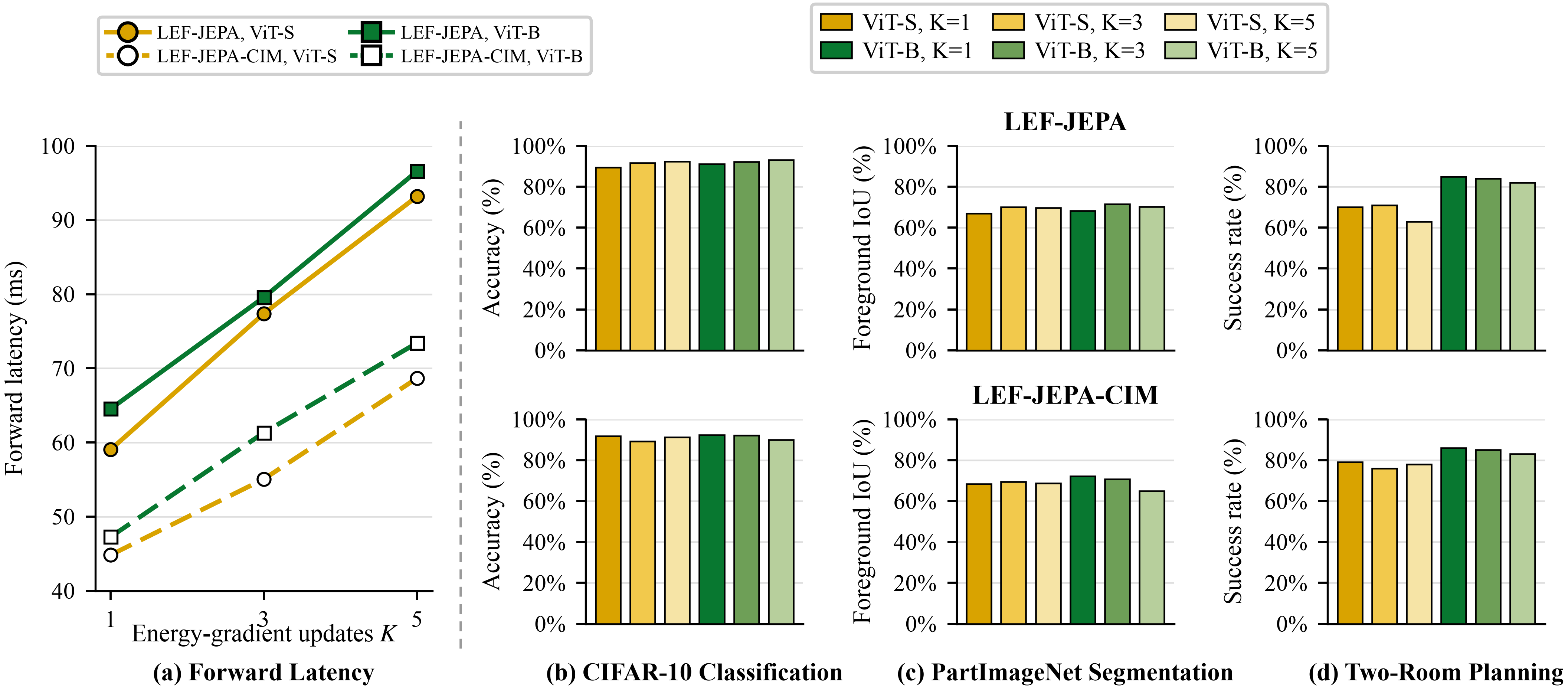}
  \caption{Effect of the number of energy-gradient updates $K$ on pretraining
  latency and downstream transfer. Each CIM recovery trajectory uses one
  physical solve and reuses its low-energy state pool across all updates. CIM
  latency excludes cloud communication and queueing; bars report means over
  three independently pretrained encoders.}
  \label{fig:supp-k-step-combined}
\end{figure}

Figure~\ref{fig:supp-k-step-combined} and Table~\ref{tab:supp-k-step} examine
how the pretraining denoising horizon $K\in\{1,3,5\}$ affects pretraining
latency and downstream transfer. The main experiments use $K=3$ for LEF-JEPA
and $K=1$ for LEF-JEPA-CIM. Because the energy module is removed after
pretraining, $K$ does not affect deployment latency. For LEF-JEPA-CIM, each
recovery trajectory queries the physical solver only once at $q_0$ and reuses
the returned state pool in subsequent updates.

\makeatletter
\setlength{\@dblfptop}{0pt}
\setlength{\@dblfpsep}{14pt}
\setlength{\@dblfpbot}{0pt plus 1fil}
\makeatother

For continuous-energy LEF-JEPA, $K=3$ gives the
highest foreground IoU on both backbones. Increasing the horizon to $K=5$
further improves classification but does not further improve segmentation or
planning performance.

The physical-Ising realization favors a shallower update. At $K=1$, ViT-B
achieves its best result on all three tasks; for ViT-S, classification and
planning are also strongest at $K=1$, while foreground IoU peaks at $K=3$.
Later updates repeatedly rescore the same state pool rather than invoking a
new physical search, which is consistent with the absence of systematic gains
beyond the first update. The measured $K=1$ forward latency is approximately
$45$--$47$ ms, below neural LEF-JEPA at $K=3$.

Bold in Table~\ref{tab:supp-k-step} marks the best depth for each method,
backbone, and task.

\begin{table}[H]
  \centering
  \caption{Complete ablation over the number of energy-gradient updates.
  Classification and foreground IoU are mean $\pm$ standard deviation across
  three independently pretrained encoders (\%); each encoder is evaluated on
  100 Two-Room episodes, and SR is averaged across the three encoder runs and
  reported to the nearest percentage point (\%).}
  \label{tab:supp-k-step}
  \scriptsize
  \setlength{\tabcolsep}{2.5pt}
  \begin{tabular*}{\textwidth}{@{\extracolsep{\fill}}llccccccccc@{}}
    \toprule
    \multirow{2}{*}{Method} & \multirow{2}{*}{Backbone}
      & \multicolumn{3}{c}{$K=1$}
      & \multicolumn{3}{c}{$K=3$}
      & \multicolumn{3}{c}{$K=5$} \\
    \cmidrule(lr){3-5}\cmidrule(lr){6-8}\cmidrule(lr){9-11}
    & & Acc. $\uparrow$ & Fg. IoU $\uparrow$ & SR $\uparrow$
      & Acc. $\uparrow$ & Fg. IoU $\uparrow$ & SR $\uparrow$
      & Acc. $\uparrow$ & Fg. IoU $\uparrow$ & SR $\uparrow$ \\
    \midrule
    LEF-JEPA & ViT-S
      & $89.40\!\pm\!.025$ & $66.88\!\pm\!.043$ & $70$
      & $91.63\!\pm\!.016$ & $\boldsymbol{70.04\!\pm\!.040}$ & $\boldsymbol{71}$
      & $\boldsymbol{92.38\!\pm\!.004}$ & $69.71\!\pm\!.024$ & $63$ \\
    LEF-JEPA & ViT-B
      & $91.00\!\pm\!.033$ & $68.28\!\pm\!.021$ & $\boldsymbol{85}$
      & $92.17\!\pm\!.009$ & $\boldsymbol{71.44\!\pm\!.021}$ & $84$
      & $\boldsymbol{93.05\!\pm\!.020}$ & $70.21\!\pm\!.035$ & $82$ \\
    \midrule
    LEF-JEPA-CIM & ViT-S
      & $\boldsymbol{91.80\!\pm\!.005}$ & $68.44\!\pm\!.039$ & $\boldsymbol{79}$
      & $89.17\!\pm\!.009$ & $\boldsymbol{69.44\!\pm\!.011}$ & $76$
      & $91.15\!\pm\!.009$ & $68.71\!\pm\!.027$ & $78$ \\
    LEF-JEPA-CIM & ViT-B
      & $\boldsymbol{92.27\!\pm\!.010}$ & $\boldsymbol{72.11\!\pm\!.044}$ & $\boldsymbol{86}$
      & $92.20\!\pm\!.004$ & $70.76\!\pm\!.021$ & $85$
      & $89.89\!\pm\!.036$ & $64.94\!\pm\!.075$ & $83$ \\
    \bottomrule
  \end{tabular*}
\end{table}

\end{document}